\documentclass[letterpaper, 10 pt, conference]{ieeeconf}

\IEEEoverridecommandlockouts 

\usepackage[binary-units=true,detect-all]{siunitx}
\usepackage[caption=false,font=footnotesize]{subfig}
\usepackage{graphicx}
\usepackage{amsmath}
\usepackage{amssymb}
\usepackage{scalerel}

\usepackage{hyperref}
\hypersetup{
  pdfinfo={
    Title={Utilizing Bluetooth Low Energy to recognize proximity, touch and humans},
    Author={Marcus M. Scheunemann},
    Subject={25th IEEE International Symposium on Robot and Human Interactive Communication, RO-MAN 2016, New York, NY, USA, August 26-31, 2016},
    Keywords={human-robot interaction, Bluetooth Low Energy, BLE, minimal hardware, robot design, proxemics, proximity, touch},
    Lang = {en-US}
  }
}

\title{\LARGE \bf
Utilizing Bluetooth Low Energy to recognize\\proximity, touch and humans
}

\author{Marcus M. Scheunemann$^{1}$, Kerstin Dautenhahn$^{1,2}$, Maha Salem$^{1}$, and Ben Robins$^{1}$
\thanks{$^{1}$All authors are with Adaptive Systems Research Group,
        University of Hertfordshire, AL10 9AB, UK
        {\tt\small \href{mailto:marcus@mms.ai}{marcus@mms.ai}}}%
\thanks{$^{2}$IEEE Senior Member}%
}

\begin{document}

\maketitle
\thispagestyle{empty}
\pagestyle{empty}

\begin{abstract}
Interacting with humans is one of the main challenges for mobile robots in a human inhabited environment.  
To enable adaptive behavior, a robot needs to recognize touch gestures
and/or the proximity to interacting individuals. Moreover, a robot
interacting with two or more humans usually needs to distinguish between
them.
However, this remains both a configuration and cost intensive task.
In this paper we utilize inexpensive Bluetooth Low Energy (BLE) devices and propose an easy and configurable technique to enhance the robot's capabilities to interact with surrounding people.
In a noisy laboratory setting, a mobile spherical robot is utilized in three proof-of-concept experiments of the proposed system architecture.
Firstly, we enhance the robot with proximity information about the individuals in the surrounding environment. Secondly, we exploit BLE to utilize it as a touch sensor. And lastly, we use BLE to distinguish between interacting individuals. 
Results show that observing the raw received signal strength (RSS) between BLE devices already enhances the robot's interaction capabilities and that the provided infrastructure can be facilitated to enable adaptive behavior in the future. We show one and the same sensor system can be used to detect different types of information relevant in human-robot interaction (HRI) experiments.
\end{abstract}

\section{INTRODUCTION}
\label{sec:intro}
Interacting with humans is one of the main tasks for mobile robots used in a human-inhabited environment and it remains an active area of research.

Our main motivation behind this work is the need of an adaptive robot to interact with children with autism for therapy purposes.
Promising results were shown when robots are utilized as mediators for therapy purposes~\cite{DiehlSchmittEtAl-12, ScassellatiAdmoniEtAl-12} or, more recently, as a diagnostic tool~\cite{BoccanfusoBarneyEtAl-16} for children and toddlers with autism.
However, long-lasting sustainable interactions or play sessions are rare.
To achieve such sustainable play sessions, we expect that the robot is capable of adapting its behavior to that of the interacting child.

To further research for robot-mediated therapy for children with autism, we therefore suggest that a robot needs to be capable of (A) being aware of individuals in its vicinity, (B) recognize touch, and (C) being able to distinguish between interacting individuals.

In general, environmental considerations are important for human-robot interaction (HRI).
Environments in experimental settings are usually designed as natural and familiar as possible for participants. This is particularly important when working with children, especially with children with autism, whereby studies are usually conducted in school environments. 

However, most state-of-the-art systems fall short in perceiving human interaction data in an unobtrusive way.
When using tracking systems and in-built cameras, the environment usually needs to fulfill certain constraints so that algorithms can be applied to distinguish between humans.
Moreover, a robot platform does not necessarily support easy expansion capabilities for attaching a camera or other tracking devices. 
In addition, mounting a camera on some platforms (e.g. spherical, rotating, mobile robot) and collecting useful information is a challenging, if not impossible task.

To engage children and toddlers in natural, spontaneous play scenarios with a robot, ideally, a non-distracting environment (i.e. familiar surroundings) should be chosen,
otherwise children's attention is easily drawn to the unfamiliar devices (e.g. cameras, tracking devices), their behavior may change due to distraction, or they may simply feel uncomfortable.
Almost always, a tracking or camera system changes the experimental environment drastically, making a long lasting familiarization phase necessary to decrease distraction.
Hiding such systems in an everyday environment, i.e. small room in a school, is rarely applicable and will be costly and time-consuming while being of uncertain outcome.

Moreover, configuration/calibration is needed when utilizing visual systems to enhance the robot's real-time capabilities. 
Participants either need to wear specific clothing in a defined location, or their face/skeleton/marker needs to be calibrated for the system.
From our experience, children who feel uncomfortable with particular items of clothing will try to remove them, try to change the configuration or simply feel distracted/uncomfortable.
In addition, it is very challenging to motivate the children to stand still during a calibration.
Thus, the state of the art capabilities of built-in cameras and external tracking systems are very often not applicable for experiments with children in naturalistic, e.g. school, environments. This, in particular, holds true for children with special needs.

In this paper we utilize Bluetooth Low Energy (BLE) devices and propose an easily applicable (i.e. with little environmental constraints), unobtrusive and inexpensive sensor system not requiring particular configuration to enable and enhance interaction between a robot and one or more individuals.
Three proof-of-concept evaluations show that our system can put the three aforementioned requirements into practice.  
Moreover, our system is applicable to almost any robotic platform without the need to expand the built-in board or re-program the firmware.

\section{BLUETOOTH LOW ENERGY}
Bluetooth Classic is widely used to pair with peripheral devices (e.g. head phones, mobile phones) to initate a data stream. Measurments of the received signal strength (RSS) has also been utilized for tasks related to this work. Proximity data to people wearing Bluetooth devices were used to infer social networks~\cite{DoGatica-Perez-11}, analyze interaction patterns with augmented objects~\cite{SiegemundFlorkemeier-03}, or to localize them indoor~\cite{SubhanHasbullahEtAl-11}.
However, Bluetooth Classic has significant shortcomings for our intended application. 
Scanning for other Bluetooth devices can take up to \SI{10.24}{\second} making it inapplicable in highly dynamic environments. Moreover, due to the protocol's focus on communication (i.e. allowing large data payload), the high energy consumption makes it unsuitable for embedded systems or to equip children unobtrusively.

In 2011 the core specification Bluetooth 4.0 was introduced~\cite{bluetooth}. Its subsystem Bluetooth Low Energy (BLE), also referred to as Bluetooth Smart, addresses these issues.
BLE utilizes \SI{2}{\mega\hertz} bands over the unlicensed \SI{2.4}{\giga\hertz} radio band.
For advertising (i.e. broadcasting), only three channels are used. Those are chosen to not collide with the most commonly used WiFi channels~\cite{FaragherHarle-15}.
BLE uses very short duration messages with small payloads, yielding low power consumption \cite{FaragherHarle-15}. Devices operating on a coin cell battery with \SI{250}{mAh} can last 1--2 years~\cite{bluetooth, RaultEtAl-14}.
Moreover, the maximum scanning time is decreased from \SI{10.24}{\second} (Bluetooth Classic) to to less than \SI{10}{\milli\second}~\cite{FaragherHarle-15}.

BLE devices are small enough to attach them to people in an unobtrusive way (see Fig.~\ref{fig:bluetooth_components}).
Many manufacturers are utilizing BLE for their devices, such as mobile phones, modern smart watches and fitness bracelets.
Therefore it can be safely assumed that BLE devices will enrich our environment in the future~\cite{FaragherHarle-15}.
Moreover, the inexpensive availability (prices may even fall due to increasing popularity) and the technological characteristics make it ideal for our application.
The widespread use of BLE in existing devices and the many upcoming additional applications and hardware devices further enhance the motivation to utilize this technology.

Although BLE is a relatively new protocol, research already facilitates the received signal strength (RSS) of BLE.
A typical scenario to utilize wireless sensor networks (WSN) such as WiFi or Bluetooth is for indoor localization purposes. A mobile receiving device collects RSS of surrounding static sensors. A map of fingerprints (each cell contains RSS data to all surrounding sensors) is computed. With this preprocessed map, an indoor position of a mobile sensor can then be inferred. 
In \cite{FaragherHarle-15}, BLE beacons are facilitated as a sensors network. Localization with BLE was shown to be more accurate compared to established WiFi localization.

In \cite{SchwarzEtAl-15}, stationary BLE devices are provided in a household environment. An object of interest (e.g. key ring) is equipped with a BLE beacon. A robot uses the RSS between the stationary devices and the beacon to infer a position of the object in question.
The following section will describe the underlying hardware used in our experiments in more detail.

\section{SYSTEM ARCHITECTURE}
\begin{figure}[htpb]
\centering
\includegraphics{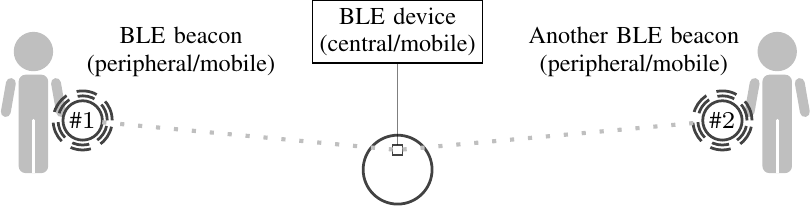}
\caption{This figure depicts the overall experimental setup. A central BLE device is used to passively scan for advertisements of peripheral BLE devices. People may be equipped with one or more advertising BLE devices. The central device computes packages with RSS data and ID of received advertisements.
These packages are provided via a wired connection to another system on the robot's board or wireless via Bluetooth connection to another machine. Since we do not have access to the robot's board, we used the Bluetooth connection yielding additional latency.}
\label{fig:concept}
\end{figure}

Figure~\ref{fig:concept} visualizes the overall setup. 
A mobile, rotating robot~(section~\ref{sec:queball}) is equipped with a central BLE device~(section~\ref{sec:BLED112}), used to constantly scan for advertisement packages of peripheral BLE devices~(section~\ref{sec:gimbal}).
Humans involved in interaction with the robot may be equipped with one or more advertising BLE devices.

The received signal strength (RSS) is computed for all packages received.
We are particularly interested in the RSS between our advertising peripheral devices to the central one on the robot.
Thus, with each received advertisement, the central device collect the corresponding RSS data and ID of the corresponding peripherals.
This information is then provided via a wired connection to another system on the robot's board or wireless with a Bluetooth connection.
Since we do not have access to the robot's board~(\ref{sec:BLED112}), we facilitate a Bluetooth connection.
The following subsections discuss the details of the system components (i.e. the robot, the BLE beacons and the BLE central device).

\subsection{The robot platform (QueBall)}
\label{sec:queball}
\begin{figure}[thpb]
\centering
\subfloat[The robot platform]{\includegraphics{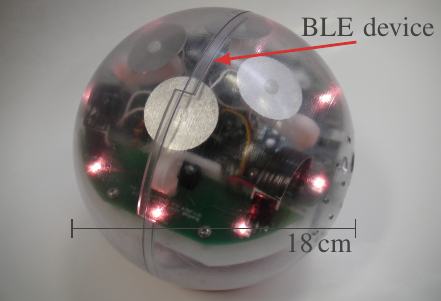}
\label{fig:queball}}
\hspace*{\fill}
\hspace{-0.13cm}
\subfloat[BLE components]{\includegraphics{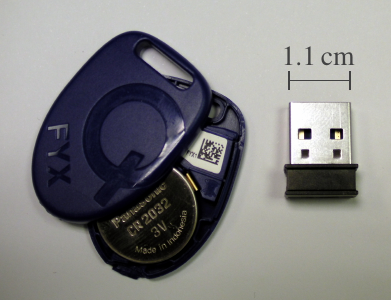}
\label{fig:bluetooth_components}}
\caption{(a) The mobile robot platform QueBall is capable of moving back/forth and tilting left/right. It emits sound, colors and detects touch (the four circle objects on top are the touch sensors). 
(b)~Self-powered and configurable advertising beacon meant to be attached to people (left); a central Bluetooth device to equip the robot to scan for surrounding peripherals signal strength (right).}
\label{fig_sim}
\end{figure}
Figure~\ref{fig:queball} depicts the spherical robot QueBall. It is developed for play sessions for children with autism and preliminary results were ``very positive''~\cite{SalterDaveyEtAl-14}. 
The robot has two degrees of freedom (DoF) through two servo motors (one servo controls back and forward motion, the other the left and right tilt).
For further control, the robot has an accelerometer and four capacitive touch sensors on the chassis pointing upwards.

The embedded robot firmware is proprietary (i.e. we do not have the possibility of changing the software), which influences the hardware decision of the BLE component (section~\ref{sec:BLED112}).

\subsection{A central scanning device Bluegiga's BLED112}
\label{sec:BLED112}
The robot is equipped with a central device to scan the environment for other BLE devices (e.g. BLE beacons). 
We choose Bluegiga's module BLED112. It is based on the \SI{6x6}{\milli\metre} CC2540F128 from Texas Instruments, a ``cost-effective, low-power, true system-on-chip (SoC)'' for BLE applications~\cite{TexasInstruments-10}. 
Hardware details can be found in the provided datasheets~\cite{TexasInstruments-10, SiliconLaboratories-15}.

The module is depicted on the right site of Figure~\ref{fig:bluetooth_components}.
Bluegiga provides a protocol BGAPI to control the integrated BLE stack (i.e. sending commands and receiving events/responses) in either one of the following two options: 
\begin{enumerate}
\item An external microcontroller/PC connected via UART or an USB port controls the stack with BGLib, an ANSI C implementation of BGAPI.
\item Facilitate the embedded CC2540 chip. Control it with BGScript, the BGAPI implementation in a \emph{Basic}-like scripting language.
\end{enumerate}
We were particularly interested in the possibility of facilitating the chip as an independent unit, since our (robot) platform does not allow us any customization of its firmware/computational unit (cp.~\ref{sec:queball}).
The central BLE device is set in a constant passive scanning mode and the machine running the robot client is connected via Bluetooth. 
Each received advertisement package triggers a response event on the chip:
\begin{enumerate}
\item Extract the sender ID and the corresponding signal strength to the sender.
\item Communicate the information to the robot client.
\end{enumerate}
Thus, signal strength changes can influence the robot's behavior. Details of the implementation can be found on our online project page\footnote{Project page: \url{http://mms.ai/BLE4HRI}}.

\subsection{Peripheral proximity beacons Gimbal Series 10}
\label{sec:gimbal}
As for the peripheral devices, we utilize small, inexpensive and configurable Gimbal Proximity Beacons (Series~10). Figure~\ref{fig:bluetooth_components} depicts the beacon. As can be seen, the beacons are powered with standard coin cell batteries.
These beacons are non-connectable and only meant to frequently advertise (i.e. broadcast) little payload so the central unit (cp.~\ref{sec:BLED112}) can measure the corresponding signal strength and infer their position. 

We used the beacons with the iBeacon protocol~\cite{Apple-14}.
Each beacon has to get assigned a \SI{20}{byte} payload. The first bits are shared among all our beacons, the last four bits specify to whom the beacon belongs and the place where the beacon is attached to the child (e.g. to the foot or wrist). 
The configuration of an iBeacon can be changed in accordance to the application needs. We set an omnidirectional antenna (instead of a directed one since we cannot influence the direction of the beacons), a maximal transmission interval of \SI{\sim100}{\milli\second} and a maximum Transmission Power (txPower) of \SI{0}{dBm}.

\section{PROOF-of-CONCEPT EVALUATION}
In this section, we present our proof-of-concept evaluation of the setting presented in the previous section. RSS data is only measured between beacons and the central device mounted on the robot. 

In the experiment in section~\ref{sec:awareness}, we study distance information between the center and peripheral BLE device.
We use a GoPro camera attached to the ceiling to collect video data in order to have a record of the experiment.
To track the robot in the video footage and collect distance information of the robot to the beacon, we used the software Kinovea\footnote{Kinovea is a free and open source solution for (live) tracking of movements in the context of sport analysis. Its simplicity seemed adequate for a proof-of-concept evaluation. Tracking was only possible in a 2D plain, experiments were setup accordingly to minimize errors.}.

In the experiments \ref{sec:touch} and \ref{sec:distinguish} a simple game was used to test the setting with a spherical, rotating, mobile robot (cp.~\ref{sec:queball}). 
Participants are equipped with wristbands with a zipper pocket. Each wristband carries one BLE beacon.
The robot moves around and changes its direction either when touched, or when moving against an obstacle. Depending on which touch sensor is triggered most, it changes its direction in the opposite direction.

We used the robot QueBall described in section~\ref{sec:queball}. The BLE central device is placed in the center of all four touch sensors (cp.~Figure~\ref{fig:queball}). 

The experiment took place in a highly noisy environment (i.e. more than 5 strong WiFi hotspots operating close-by) to check whether this technology is applicable in an everyday setting.  

\subsection{Proximity}
\label{sec:awareness}
\begin{figure}[htpb]
\centering
\includegraphics[width=\columnwidth]{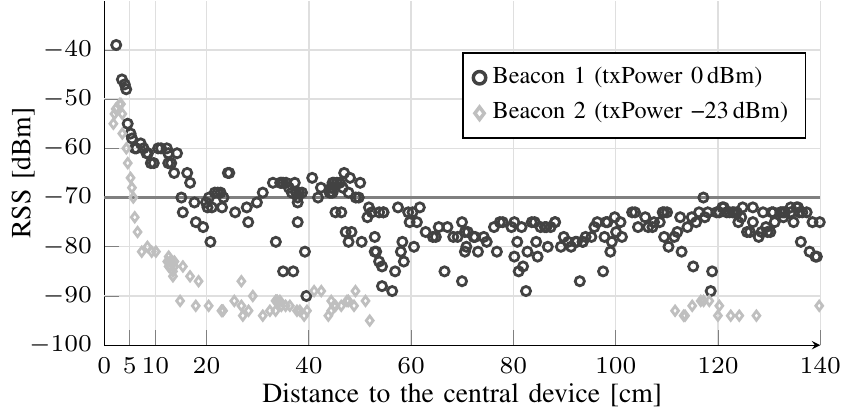}
\caption{Beacon 1 was set up with the highest possible transmission power (i.e. txPower) of \SI{0}{dBm} (highest) and beacon 2 to the lowest \SI{-23}{dBm}. While a beacon was moved with roughly constant speed of \SI[per-mode=fraction]{\sim5}{\centi\metre\per\second} away from the central device, the distance between beacon and central device was tracked.
The graph depicts the RSS value of each beacon. It shows that the measurements may yield information about the proximity of a person.}
\label{fig:comparedBm}
\end{figure}%
The iBeacon protocol is meant to be capable of distinguishing three different proximity states: immediate (few centimeters), near (1--3 meters), far (none of the aforementioned)~\cite{Apple-14}.
However, the manufacturer states that ``it is critical to perform calibration''~\cite{Apple-14}. 
The calibration described in the manual needs a beacon deployed to a known position.
Since all BLE devices in our setting are mobile, this is not applicable.
In this experiment, we investigate whether RSS data can be used to infer proximity information, although none of the BLE devices involved has a fixed position.
In Figure~\ref{fig:comparedBm}, beacon 1 was set to the highest possible transmission power of \SI{0}{dBm} (black), beacon 2 was set to the lowest possible transmission power \SI{-23}{dBm} (gray). In two consecutive sessions, each beacon was manually moved away from the central device with a speed of \SI[per-mode=fraction]{\sim5}{\centi\metre\per\second}.

As we discussed earlier, partial occlusion with a human hand or body and other sensor noise may yield a very low RSS value, despite the beacon being very close to the robot. Without any additional filtering methods or additional assumptions, it is challenging to derive the precise human distance to the robot just by its signal strength.

However, one can infer that a high RSS value implies physical closeness. This is due to the lower possibility of a beacon being occluded when closer to the receiving device. In addition, other radio noise interferes less since the signal does not have to travel a long distance. Thus, strong signal false positives are less expected~\cite{FaragherHarle-15, SchwarzEtAl-15}.

We implemented a state machine which assumes that a human is in the robot's vicinity (i.e. in the same room) whenever his beacon's RSS is present.
The signal must be confirmed within a couple of seconds, otherwise the human is considered to be out of the vicinity of the robot.
Already this method extends the spectrum of the robot's activities (e.g. behave differently when more than one child is present).

A threshold-based method is applicable for measuring whether a human is ``close''. For the given example, a RSS of \SI{-70}{dBm} is a good enough threshold to assume the corresponding human in an environment of \SI{<50}{\centi\metre}.
In practice (i.e. play session), this threshold varies in accordance to the direction of the antennas and motion of both BLE devices.

For our first trials, we found that \SI{-60}{dBm} is a good enough threshold for the state machine to consider the beacon within the room, but not closer than \SI{10}{\centi\metre} when RSS values are lower than \SI{-60}{dBm}. Further research is needed to verify this observed threshold. 

Thus, we were able to distinguish between three different ranges: 1) Close, 2) in the same room, and 3) outside the room.
Although this observation is encouraging, further research is needed to verify this capability for other environments.
To have the same effect in a general setting during a play session with children, one can think of equipping the children with the beacon only when they enter the room as part of the game.
\subsection{Touching}
\label{sec:touch}
With the following experiments, we investigate whether a BLE device can be utilized as a touch sensor.
We choose to take the perceived touch of the robot as ground truth data.
Our goal is not to replace the robot's own touch sensors, but to show that our proposed multi-purpose sensor system can yield information about touch.

As described, the central BLE device is in the center of the four touch sensors of the robot.
Whenever at least one of the four touch sensors is triggered, we consider it a touch gesture.
This information provides us with ground truth data when the robot is actually being touched.

\begin{figure}[htpb]
\centering
\includegraphics[width=\columnwidth,page=1]{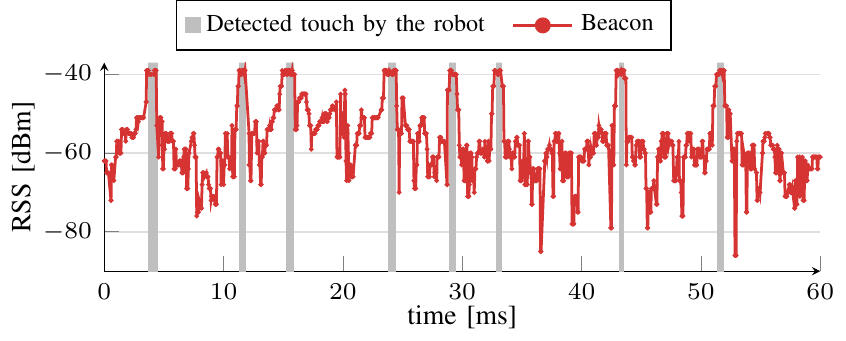}
\caption{The gray bars indicate when at least one touch sensor is triggered. The signal strength between a BLE device and the robot's central BLE device is visualized in red. The graph shows an \SI{60}{\second} excerpt of the experiment. It seems likely that a person equipped with a BLE beacon can deliberately control a robot with touch gestures. The person was aware of the technology involved. Over a \SI{5}{\minute} experiment, most values greater than \SI{-41}{dBm} corresponds to the touch perceived by the robot's touch sensors.}%
\label{fig:delibTouch}
\end{figure}%

We conducted a \SI{10}{\minute} experiment with a moving robot and one participant. 
We used the game described in the introduction and recorded each frame the robot sends (roughly every \SI{50}{\milli\second}).
The participant is well aware of the position of the sensor and receiver. It is investigated whether he is able to interact with the robot in such a way, that his touches are recognized.

As discussed earlier, it is safe to assume that only a very close beacon emits a strong signal strength. 
Figure~\ref{fig:delibTouch} shows an excerpt (\SI{60}{\second}) of the experiment with a human who wants to interact with the robot deliberately.

\begin{table}[htpb]
\renewcommand{\arraystretch}{1.3} 
\caption{Analyze received frames} 
\label{tbl:touch} 
\centering 
\begin{tabular}{|c|c|c|} 
\hline
\bfseries RSS value & \bfseries Touch occurs & \bfseries  No touch occurs \\
\bfseries greater than: & Total: 413 & Total: 5277 \\
\hline\hline 
\bfseries \SI{-40}{dBm} & 214 (\SI{52}{\percent}) & 244 (\SI{5}{\percent}) \\ 
\hline 
\bfseries \SI{-41}{dBm} & 389 (\SI{94}{\percent}) & 370 (\SI{7}{\percent}) \\ 
\hline 
\bfseries \SI{-42}{dBm} & 396 (\SI{96}{\percent}) & 404 (\SI{8}{\percent}) \\ 
\hline 
\end{tabular} 
\end{table}

To recognize whether a touch gesture occurs, a RSS-threshold filter can be applied.
Table~\ref{tbl:touch} shows the result of the overall (\SI{10}{\minute}) experiment.
Frames received when the person touches the robot are analyzed checking whether the belonging RSS value is greater than either \SI{-40}{dBm}, \SI{-41}{dBm}, or \SI{-42}{dBm}, i.e., whether the given threshold indicates a touch gesture.
It is also analyzed how often the same threshold corresponds to a frame when no touch is conducted.

The threshold of \SI{-41}{dBm} is a promising choice. 
The RSS value of most frames corresponding to a touch gesture is \SI{94}{\percent}. The filter however gives a \SI{7}{\percent} confirmation of touch although no touch sensor is registered.
It is worth noting that all these frames are in the direct vicinity (\SI{\pm400}{\milli\second}) of a touch gesture.

This experiment shows that BLE can be used to enable touch capabilities of a robot.
A human who is introduced to the technology is able to control the robot with touch input very accurately.
However, whenever humans are not aware of how to exactly to facilitate the BLE devices (e.g. when used in a game scenario with children), it remains challenging to infer a touch gesture solely by the raw RSS input.
As an example, one can think of a person positioning his hand very close over the robot's touch sensor without touching.
Thus, the RSS value may be frequently very high. The proposed filter will consider this as a touch gesture.
We assume that the fusion of other sensor input (e.g. an accelerometer) expands the possibilities of distinguishing a touch gesture from the given example.

This results show that BLE can be used to enable touching capabilities for a robot. Further investigation may find more advanced filters which do not need a fixed threshold. 
Since the used game was coupled to the built-in touch sensors of the robot, it is worth noticing that the results may be strongly affected by characteristics of these sensors.
Future research may be conducted to examine whether a game only relying on the raw RSS data (i.e. without using the built-in touch sensors) may yield a acceptable game play.
\subsection{Distinguishing between humans}
\label{sec:distinguish}
\begin{figure}[htpb]
\centering
\vspace*{.25cm}\includegraphics{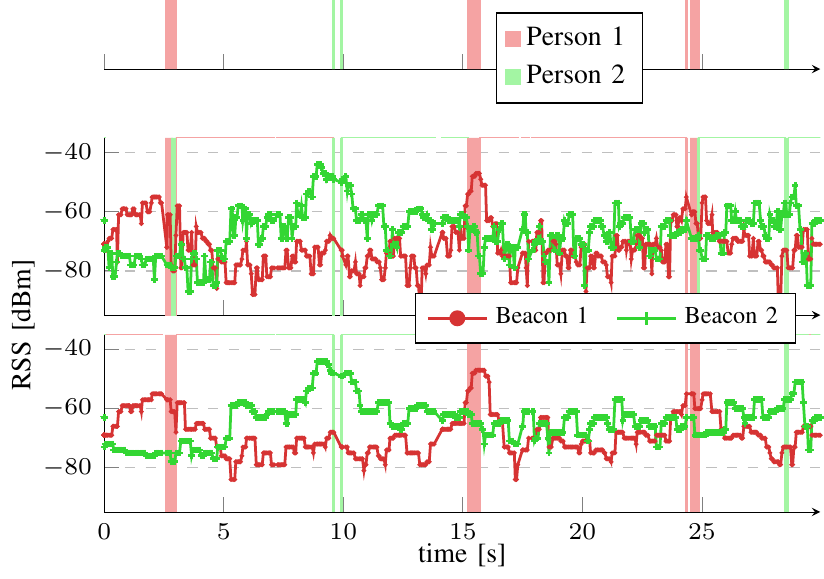}
\caption{Two participants controlling a moving robot by touch gestures.
The areas in the top graph indicate that one of the touch sensors of the robot was triggered. These ground truth information of touch is provided by the in-built touch sensors of the robot. 
The areas are colored in accordance to the person who touched the robot (red: person 1; green: person 2).
The associations are received by observation of the video footage.
The red and the green lines depict the RSS to the person's beacons (middle/bottom).
The middle graph depicts the raw signal. When the robot is touched, a person is associated with the touching action who's beacon emits the highest signal strength (areas are colored accordingly). The 1\textsuperscript{st} and the 4\textsuperscript{th} touch sequence is associated wrongly (partially).
The bottom graph depicts the smoothed RSS. A sliding window of \SI{300}{\milli\second} is used. Thus, each time an action is computed, only the maximum RSS value of the last \SI{300}{\milli\second} is considered and the touch is associated correctly.
}
\label{fig:twoHumans}
\end{figure}%
Our major concern was to enhance the level of adaptability in a play session. In other words, a robot should be capable of distinguishing between two or more children. 
Section~\ref{sec:touch} shows that there is no link between RSS and distance, since partial occlusion during a touch yields a low signal strength. Since we intend to use the setting in a play session with children, we may not be able to explain how they should touch the robot. Instead, whenever a touch is triggered, we want to infer the play partner.

We conducted a \SI{510}{\second} experiment with two participants playing the game described in the introduction. Thus, the robot was constantly moving and changing its direction according to touch input.
They were not given any detailed explanations of the wristband and sensor, except that it was required to touch the robot with the hand wearing the wristband.
The robot sends information roughly every \SI{50}{\milli\second}. Each beacon is meant to advertise every \SI{100}{\milli\second}.
For each received robot message, we compute the person who was touching the robot.
Figure~\ref{fig:twoHumans} depicts an excerpt of this experiment. The time when either one of the robot's touch sensors is triggered is depicted as a red or green area, depending on whom touches (red: person 1; green: person 2). The measurements of two different wristbands are depicted in red and green solid lines.

As discussed earlier, there is no fixed threshold to determine whether a touch is triggered.
Moreover, Figure~\ref{fig:twoHumans} (middle) depicts partial occlusion during a touching sequence (see 1\textsuperscript{st} and 4\textsuperscript{th} touch).

As a first approach, we decided to concentrate on the relative maximum when a touch is triggered. This means, instead of comparing total values, the wristband emitting the highest RSS during a touch is considered to be the wristband belonging to the touching person.
\SI{95}{\percent} of the frames were associated correctly (cp.~table~\ref{tbl:distinguish}). We also studied whether a touch sequence was associated correctly. This holds true for \SI{90}{\percent} of all touch sequences.

\begin{table}[htpb] 
\renewcommand{\arraystretch}{1.3} 
\caption{}%
\label{tbl:distinguish} 
\centering 
\begin{tabular}{|c||c|c||c|c|} 
\hline
							 & \multicolumn{2}{c||}{\bfseries Received frames} & \multicolumn{2}{c|}{\bfseries  Touch sequences} \\
							 & \multicolumn{2}{c||}{Total: 613} & \multicolumn{2}{c|}{Total: 98} \\
\hline 
\bfseries Window & \bfseries Positive & \bfseries False & \bfseries Positive & \bfseries False\\ 
\hline\hline 
\SI{0}{\milli\second} & 584 (\SI{95}{\percent}) & 29 & 88 (\SI{90}{\percent}) & 10\\ 
\SI{300}{\milli\second} & 599 (\SI{98}{\percent}) & 14 & 93 (\SI{95}{\percent}) & 5\\ 
\SI{500}{\milli\second} & 600 (\SI{98}{\percent}) & 13 & 94 (\SI{96}{\percent})& 4\\ 
\hline 
\end{tabular} 
\end{table}

Since it is unlikely that RSS becomes underestimated (lesser occlusion possibilities with smaller distance),
we compute the maximimum RSS value of each wristband of a previous \SI{300}{\milli\second} window. The resulting graph is depicted in Figure~\ref{fig:twoHumans} (bottom).
The touch is associated with the correct person, as depicted with red and green areas.
For the whole experiment, a window of \SI{300}{\milli\second} (\SI{500}{\milli\second}) will distinguish \SI{95}{\percent} (\SI{96}{\percent}) of all touch events correctly (cp.~table~\ref{tbl:distinguish}).

This experiment shows that the technology can be used to distinguish between interacting humans without any configuration involved.
Further research will increase correctly associated messages and touch sequences. For example, one can take into account that the beginning of a touch is most likely the less occluded measurement of the touching person.  

\section{FURTHER WORK}
Among already addressed ideas, the given architecture may be tested with other beacons and protocols. For example, more expensive beacons such as eddystone beacons allow a higher transmission rate than the proprietary iBeacon protocol, which may increase the accuracy of the underlying RSS measurmenets~\cite{FaragherHarle-15}. 
Other proximity protocols may provide information about the channel used for advertisement. The iBeacon protocol has no such information and thus, a smearing effect through merging all measurments of the three advertisment channels is present~\cite{FaragherHarle-15}.

Naturally, more BLE devices, filters, or integrating other sensors may enhance the information accuracy. 
For example, equipping the robot with an inertial measurement unit (IMU) will allow us to derive the orientation of the BLE central device. An increase of a signal strength value in accordance to a specific direction may enable the robot to be able to approach or avoid an interacting individual. 
Moreover, having a BLE device also on the back of an individual may ease the process of dealing with occlusions.

Most important, the technology enables a broad range of robot behaviors.
The greater variety of behaviors will increase the adaptability of a robot, which is hoped to result in more sustainable human-robot interactions with autistic children during play sessions.

\section{CONCLUSION}
In this paper, we exploited inexpensive, off-the shelf, low power-consuming, and easily applicable BLE components to enhance HRI scenarios.
The approach yields little environmental distraction and only few configuration needs.
A central BLE device was attached to a robot, interacting participants were equipped with wristbands carrying BLE beacons.
The central device scans for advertisements of the BLE beacons, and provides the RSS data between the central and each advertising beacon to a robot controller. In this work, we concentrate on exploiting either the raw signal strength data of BLE or using the maximum RSS value of a specific time window.

Our proof-of-concept evaluations showed that one and the same low-cost technology can be used to extract different types of information factors relevant in HRI:
\renewcommand{\theenumi}{\Alph{enumi}}
\begin{enumerate}
\item Increase the robot's \emph{awareness} of participants being present in its environment/proximity,
\item using a BLE device as a touch sensor for deliberate use, 
\item and enabling the robot to distinguish between interacting individuals.
\end{enumerate}

The greatest strength of the system architecture is its flexibility. The small independent devices enable flexible positioning and provide all three information factors to a robot controller. Before mounting them physically to a robot/object/board, prototyping can be conducted easily.
Although we discussed the results only for a mobile spherical robot, initial experiments show that the technology can be easily applied to other robotic platforms or even non-robotic items (e.g. an actual ball).
For example, a toy can be equipped with the central device. The robot could adapt to the children's interaction with this toy and may encourage the child to play with it.

The widespread use of BLE in our everyday devices (e.g. mobile phones, bracelets) may offer an additional information source to robots intended to be present in human inhabited environment.
For example, these robots may use the presented system as an additional update of their beliefs inferred through their visual system.

\end{document}